\documentclass[letterpaper, 10 pt, conference]{ieeeconf}  

\IEEEoverridecommandlockouts                              

\usepackage{graphics} 
\usepackage{epsfig} 
\usepackage{mathptmx} 
\usepackage{times} 
\usepackage{amsmath} 
\usepackage{amssymb}  

\usepackage{booktabs}
\usepackage{tabulary}
\usepackage{array, multirow}
\usepackage{cite}
\usepackage{rotating}
\usepackage{xcolor}
\usepackage{xspace}

\usepackage[breaklinks=true,bookmarks=false]{hyperref}

\makeatletter
\DeclareRobustCommand\onedot{\futurelet\@let@token\@onedot}
\def\@onedot{\ifx\@let@token.\else.\null\fi\xspace}

\def\eg{\emph{e.g}\onedot} 
\def\ie{\emph{i.e}\onedot} 
\def\cf{\emph{c.f}\onedot} 
 
\def\wrt{w.r.t\onedot} 

\makeatother

\newcommand{\PAR}[1]{\vskip4pt \noindent{\bf #1~}}

\newcommand{\qj}[1]{\textcolor{red}{(#1)}\xspace}                     

\definecolor{darkgreen}{RGB}{15, 132, 54}
\definecolor{darkblue}{RGB}{15, 39, 132}
                
\newcommand{\hco}[1]{\textcolor{red}{#1}\xspace}                    
\newcommand{\hcp}[1]{\textcolor{purple}{#1}\xspace}                    

\title{\LARGE \bf
To Learn or Not to Learn: Visual Localization from Essential Matrices
}

\author{Qunjie Zhou$^1$, Torsten Sattler$^2$, Marc Pollefeys$^{3,4}$, Laura Leal-Taix\'{e}$^1$
\thanks{This research was partially funded by the
Humboldt Foundation through the Sofja Kovalevskaya Award.}%
\thanks{$^1$Technical University of Munich}%
\thanks{ $^2$Chalmers University of Technology}%
\thanks{$^3$Department of Computer Science, ETH Z\"{u}rich}%
\thanks{$^4$Microsoft Z\"{u}rich}%
}
\begin{document}

\maketitle
\thispagestyle{empty}
\pagestyle{empty}

\begin{abstract}
Visual localization is the problem of estimating a camera within a scene and a key technology for autonomous robots. 
State-of-the-art approaches for accurate visual localization use scene-specific representations, resulting in the overhead of constructing these models when applying the techniques to new scenes. Recently, learned approaches based on relative pose estimation have been proposed, carrying the promise of easily adapting to new scenes. 
However, they are currently significantly less accurate than state-of-the-art approaches. In this paper, we are interested in analyzing this behavior. To this end, we propose a novel framework for visual localization from relative poses. Using a classical feature-based approach within this framework, we show state-of-the-art performance. Replacing the classical approach with learned alternatives at various levels, we then identify the reasons for why deep learned approaches do not perform well. Based on our analysis, we make recommendations for future work.
\end{abstract}


\section{INTRODUCTION}
Given a query image, the goal of visual localization algorithms is to estimate its camera pose, \ie, the position and orientation from which the photo was taken. 
Visual localization is a fundamental step in the perception system of robots, 
\eg, autonomous vehicles~\cite{Lim12CVPR,Lynen2015RSS}, and a core technology for Augmented Reality applications~\cite{Arth09ISMAR,Castle08ISWC}.

Current approaches to visual localization that achieve state-of-the-art pose accuracy are based on 3D information~\cite{Sattler2017PAMI,Brachmann2018CVPR,Cavallari2017CVPR,Toft2018ECCV,Schoenberger2018CVPR,Meng2017IROS,Taira2018CVPR,Sarlin2019CVPR}. 
They first establish 2D-3D matches between 2D pixel positions in a query image and 3D points in the scene. 
The resulting correspondences are then used to estimate the camera pose 
~\cite{Kneip2011CVPR,Fischler81CACM}. 
The 3D scene geometry can either be represented explicitly through a 3D point cloud or implicitly via the weights of a convolutional neural network (CNN). 
Both types of representations are scene-specific, \ie, 
they need to be build per scene and do not generalize to unseen scenes. 


A more flexible scene representation models a scene through a set of database images with associated camera poses~\cite{Sattler2017CVPR}. 
Building such a scene representation is trivial as it amounts to adding posed images to a database. 
The pose of the query image can then either be approximated by the pose of the most similar database image(s)~\cite{Torii15CVPR,Arandjelovic16CVPR,Torii2011ICCVW,Zamir10ECCV}, identified through image retrieval~\cite{Sivic03ICCV,Philbin07CVPR}, or computed more accurately~\cite{Zhang06TDPVT,Taira2018CVPR,Sattler2017CVPR}.
Multiple methods based on deep learning have been proposed for estimating the pose of the query relative to the database images~\cite{Balntas2018ECCV,Laskar2017ICCVW,Melekhov2017ICAC,Ummenhofer2017CVPR,Zhou2017CVPR} rather than to compute it explicitly from feature matches~\cite{Zhang06TDPVT}. 
However, 
such approaches do not consistently perform better than a simple retrieval approach that only approximates the query pose~\cite{Sattler2019CVPR}. 

Visual localization approaches based on relative poses have desirable properties, namely simplicity and flexibility of the scene representation~\cite{Sattler2017CVPR} and easy adaption to new scenes, compared to 3D-based approaches. 
Also, leveraging modern machine learning techniques for relative pose estimation seems natural. 
This leads to the question why learning-based approaches do not perform well in this setting. 

The goal of this paper is to analyze the impact of machine learning on relative pose-based localization approaches. To this end, we propose a novel and generic framework for visual localization that uses essential matrices 
inside a novel RANSAC scheme to recover absolute poses. 
Our framework is agnostic to the way the essential matrices are estimated. 
We thus use it to analyze the impact of employing machine learning in various ways:  
We compare (a) a classical approach based on SIFT features~\cite{Lowe04IJCV} 
   to (b) directly regressing an essential matrix (using a novel CNN-based approach proposed in this paper)  and (c) a hybrid approach that uses learned feature matching instead of SIFT. 
Through detailed experiments, we show that: 
1) Our SIFT-based approach (a), despite its simplicity, is competitive with respect to significantly more complex state-of-the-art approaches~\cite{Sattler2017PAMI,Brachmann2018CVPR,Taira2018CVPR}, thus validating our framework. 
2) Our regression-based approach (b), although outperforming previous work, is still significantly worse than the SIFT-based variant. Also, it does not generalize to unseen scenes due to the inability of its regression layers to learn the general concepts underlying relative pose estimation. 
3) While the regression layer is mainly responsible for the inaccurate pose estimates of relative pose regression-based methods, it is not the only part that needs improvements. 
Rather, using features learned by such methods in our hybrid approach (c) also leads to less accurate results compared to (a). 
Besides proposing a novel localization framework, 
this paper thus also contributes important insights into future work towards truly generalizable learning-based visual localization.

\section{RELATED WORK}\label{sec:related_work}
\PAR{Feature-based localization.} Feature-based approaches to visual localization can be classified into \emph{direct}~\cite{Cao14CVPR,Choudhary12ECCV,Li10ECCV,Li12ECCV,Lynen2015RSS,Sattler2017PAMI,Svarm2017PAMI,Zeisl2015ICCV,Geppert2019ICRA} and \emph{indirect}~\cite{Arandjelovic14ACCV,Cao13CVPR,Irschara09CVPR,Torii15CVPR,Zamir14PAMI,Zhang06TDPVT,Sarlin2018CoRL, Sarlin2019CVPR} approaches. 
The former follow the strategy outlined above and obtain 2D-3D matches by directly comparing feature descriptors extracted from a query image with 3D points in the SfM model. 
While producing accurate camera pose estimates, their scalability to larger scenes is limited, partially due to memory consumption and partially due to arising ambiguities~\cite{Li12ECCV}. 
The former can be addressed by model compression~\cite{Li10ECCV,Cao14CVPR,Camposeco2019CVPR,Lynen2015RSS,Sattler2015ICCV} at the price of fewer localized images
~\cite{Lynen2015RSS,Cao14CVPR}. 

\emph{Indirect} approaches first perform an image retrieval 
step~\cite{Jegou08ECCV,Philbin07CVPR,Sivic03ICCV} against the database images used to build the SfM model. 
An accurate pose estimate can then be obtained by descriptor matching against the points visible in the top-retrieved images~\cite{Irschara09CVPR,Sattler2015ICCV,Taira2018CVPR}, which can be 
loaded from disc on demand. 
The retrieval step can be done very efficiently 
using compact image-level descriptors~\cite{Arandjelovic16CVPR,Radenovic2016ECCV,Torii15CVPR}. 
It is not strictly necessary to store a 3D scene representation for accurate pose estimation: 
Given the known poses of the database images, it is possible to compute the query pose via 
computing a local SfM model online~\cite{Sattler2017CVPR} or by triangulating the position of the query image from relative poses \wrt the database images~\cite{Zhang06TDPVT}. While \cite{Zhang06TDPVT} is limited to using only two database images, we propose a more general RANSAC-based approach to use more database images. 

\PAR{Learning for visual localization.}
%
Retrieval methods \cite{Arandjelovic16CVPR,Gronat13CVPR,Cao13CVPR,Nathan2019ICRA} have benefitted greatly from deep learning. 
For 3D structure-based localization, several works have proposed to directly learn the 2D-3D matching function~\cite{Brachmann2017CVPR,Brachmann2018CVPR,Cavallari2017CVPR,Donoser14CVPR,Guzman14CVPR,Shotton2013CVPR,Valentin15CVPR,Weinzaepfel2019CVPR}. 
%
Their main drawback, besides scaling to larger scenes~\cite{Brachmann2018CVPR,Schoenberger2017ARXIV}, is that they need to be trained specifically per scene. 
Recent work has shown the ability to adapt a model trained on one scene to new scenes on-the-fly~\cite{Cavallari2017CVPR}. 
Yet, \cite{Cavallari2017CVPR} considers the problem of re-localization against a trajectory while we consider the problem of localization from a single image.

\PAR{Learning absolute pose estimation.}
An alternative to regressing 2D-3D matches is to learn the complete localization pipeline, either via classification~\cite{Weyand2016ECCV} or camera pose regression~\cite{Kendall2015ICCV,Kendall2016ICRA,Kendall2017CVPR,Melekhov2017ICCVW,Naseer2017IROS,Walch2017ICCV}. 
These methods typically only require images and their corresponding camera poses as training data and minimize a loss on the predicted camera poses~\cite{Kendall2015ICCV,Kendall2017CVPR}. 
However, using 2D-3D matches as part of the loss function can lead to more accurate results~\cite{Kendall2017CVPR}. 
Similar to regressing 2D-3D matches, the learned representations are scene-specific and do not generalize. 
While methods that operate on individual images are not significantly more accurate than simple retrieval baselines~\cite{Sattler2019CVPR}, using image sequences for pose regression can significantly improve performance~\cite{Radwan2018RAL,Valada2018ICRA}. In this paper, we however focus on the single-image case.

\PAR{Learning relative pose estimation.} 
In contrast to absolute pose estimation, which is a scene-specific task, learning to predict the relative pose between images is a problem that should generalize to unseen scenes. 
%
\cite{Ummenhofer2017CVPR} propose a CNN that jointly predicts a depth map for one image and the relative pose \wrt a second image. 
In contrast to our approach, theirs requires depth maps for training. 
\cite{Zhou2017CVPR} 
is trained purely on a stream of images by using image synthesis as a supervisory loss function. 
The method is tested in an autonomous driving scenario that exhibits planar motion. Extending this method to the 6DOF scenario with larger baselines considered in this paper seems non-trivial. 
\cite{Balntas2018ECCV} propose a 
network is jointly trained for the tasks of image retrieval (based on a novel frustum overlap distance) and relative camera pose regression.
The latter is 
based on a $\mathbb{SE}(3)$ parameterization. 
Yet, the ability of \cite{Balntas2018ECCV} to generalize to new scenes is rather limited~\cite{Sattler2019CVPR}.

%
%
Most similar to our approach, \cite{Laskar2017ICCVW} first identifies potentially relevant database images via image retrieval. 
A CNN is then used to regress the relative poses between the query and the retrieved images, followed by triangulation to estimate the query's absolute camera pose inside a RANSAC loop. 
\cite{Laskar2017ICCVW} needs to find a weighting between the positional and rotational parts of the pose loss during training, which potentially needs to be adjusted per scene. 
We show that regressing essential matrices is a better choice. 
We also show how the resulting pose ambiguity can be handled via a novel RANSAC scheme. 
We analyze which parts of the localization pipeline fail when replaced by a data-driven approach, showing that learning the whole pipeline as in  \cite{Laskar2017ICCVW} is by far not the most accurate solution.

\section{ESSENTIAL MATRIX BASED LOCALIZATION}
\label{sec:localization}

\begin{figure}[t]
  \centering
  \includegraphics[width=0.48\textwidth]{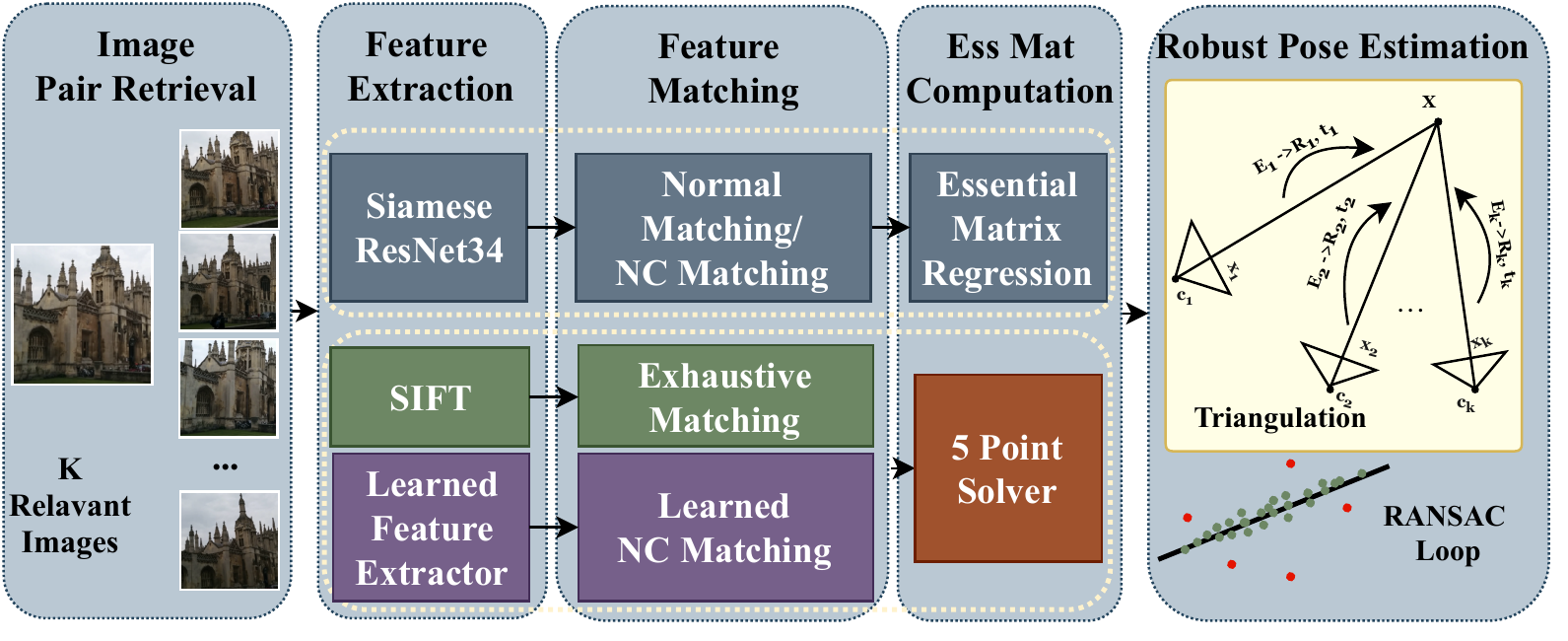}
  \caption{Our localization pipeline: The pipeline first retrieves top-k similar training images for a query image using DenseVLAD descriptors, composing k input pairs. In the next stage, one of 3 approaches(Sec.~\ref{sec:essential_estimation}) is used to estimate k essential matrices, which are fed into our RANSAC loop for relative pose extraction as well as the absolute pose computation.}
  \label{fig:pipeline}  
      \vspace{-0.5cm}
\end{figure}

In this section, we propose a scalable pipeline to estimate the absolute pose of a query image \wrt a scene represented by a database of images with known camera poses.  
Our pipeline, shown in Fig.~\ref{fig:pipeline}, consists of three stages: 
 (1) we use image retrieval to identify a set of images that potentially depict the same part of the scene as the query image (\cf Sec. \ref{sec:retrieval}). 
(2) for each retrieved image, we compute the essential matrix that encodes its relative pose \wrt the query image (\cf Sec. \ref{sec:localization:relative_pose}).
(3) using the known absolute poses of the retrieved images and the essential matrices, we 
estimate the absolute pose of the query (\cf Sec. \ref{sec:localization:ransac}).

\PAR{Why essential matrices?} 
Since we are ultimately interested in extracting relative poses, one might wonder why not training a CNN to directly predict relative poses instead of essential matrices. Several works \cite{Laskar2017ICCVW, Saha2018BMVC} propose a model for relative pose prediction, with the main disadvantage of needing a scene-dependent hyperparameter 
(\cf Sec.~\ref{sec:essential_estimation:essnet}).

We notice that directly regressing essential matrix automatically resolves the scene-dependent weighting issue from relative pose regression and also leads to more accurate results.
While directly decomposing the essential matrix into relative poses results in ambiguities, Sec. \ref{sec:localization:ransac} shows how these ambiguities can be handled inside a RANSAC loop. 

In the following, we 
describe our 3D model-free localization pipeline based on essential matrices, which is oblivious to the source of the essential matrices. 
Sec.~\ref{sec:essential_estimation} then discusses multiple approaches to essential matrix estimation.

\PAR{Notation.} The absolute camera pose ($\mathtt{R}_\mathcal{I}$, $\mathbf{t}_\mathcal{I}$) of an image $\mathcal{I}$ is defined by a rotation matrix $\mathtt{R}_\mathcal{I}$ and a translation $\mathbf{t}_\mathcal{I}$ such that  $\mathtt{R}_\mathcal{I} \mathbf{x} + \mathbf{t}_\mathcal{I}$ transforms a 3D point $\mathbf{x}$ from a global coordinate system into the local camera coordinate system of $\mathcal{I}$. 
Accordingly, the camera center $\mathbf{c}_\mathcal{I}$ of $\mathcal{I}$ in global coordinates is given by $\mathbf{c}_\mathcal{I}=-\mathtt{R}_\mathcal{I}^T \mathbf{t}_\mathcal{I}$. 
Notice that in practice, we are representing the rotation $\mathtt{R}_\mathcal{I}$ by a quaternion $\mathbf{q}_\mathcal{I}$. 
As such, we will interchangeably use either a rotation matrix $\mathtt{R}$ or a quaternion $\mathbf{q}$ to denote a (relative or absolute) rotation.

\subsection{Retrieving Relevant Database Images}  
\label{sec:retrieval}
We perform image retrieval 
by representing each database image by 4096-dimensional DenseVLAD descriptors~\cite{Torii15CVPR}, 
which has been shown to work under challenging conditions~\cite{Sattler2018CVPR}.  
Compared to other learned pipelines for image retrieval~\cite{Radenovic2016ECCV,Radenovi2018TPAMI}, DenseVLAD shows better generalization to unseen scenes, which fits well to our pipeline.


\PAR{Pair selection.} 
Simply picking top-k ranked retrieved images for each query image is not sufficient to obtain good performance. The top-k retrieved images are often taken from very similar poses, which causes problems when we estimate the camera position of the query via triangulation. 
We want to ensure larger triangulation angles 
while still keeping enough visual overlap for successful relative pose estimation. 
Starting with the top retrieved image, we thus iteratively select retrieved images to have certain minimal/maximal distances to the previously selected ones. 
The resulting query-database image pairs are then used for essential matrix estimation (\cf Sec.~\ref{sec:essential_estimation:essnet}). 
We handle outlier pairs, \ie, database images depicting a different part of the scene than the query, robustly 
via RANSAC~\cite{Fischler81CACM} as detailed in Sec.~\ref{sec:localization:ransac}.

\subsection{Pairwise Relative Pose Estimation} 
\label{sec:localization:relative_pose}
For each image pair, we compute the essential matrix $\mathtt{E}$ that encodes the relative pose between the query and database image. 
Next, we 
extract the four relative poses $(\mathtt{R}, \mathbf{t})$, $(\mathtt{R}, -\mathbf{t})$, $(\mathtt{R}', \mathbf{t})$, $(\mathtt{R}', -\mathbf{t})$    corresponding to $\mathtt{E}$~\cite{Hartley2004}, where $\mathtt{R}$ and $\mathtt{R}'$ are related by a $180^{\circ}$ rotation around the baseline~\cite{Hartley2004}. 
%
Traditionally, a \textit{cheirality test} based on feature matches is used to find the correct relative pose among the four candidates~\cite{Hartley2004}. 
However, methods that directly regress the relative pose typically do not provide such matches. 
%

We use triangulation based on the estimated relative poses and the known absolute poses of the database images to estimate the position from which the query image was taken. 
We thus only need to disambiguate the two rotations, since the position of a point triangulated from multiple directions $\mathbf{t}_1, \dots, \mathbf{t}_n$ does not change when flipping the signs of any direction $\mathbf{t}_i$. 
Thus, the absolute position of the query image can be uniquely determined by $n \geq 2$ images pairs. 
Let $\mathtt{R}_i$ and $\mathtt{R}'_i$ be the possible relative rotations that transform from the local system of $i$-th retrieved image $\mathcal{I}_i$ to the local of the query image $\mathcal{I}_q$. 
Thus, the absolute rotation part of the query image is either $\mathtt{R}_i \mathtt{R}_{\mathcal{I}_i}$ or $\mathtt{R}'_i \mathtt{R}_{\mathcal{I}_i}$.
Counting also the relative rotations estimated from another image pair ($\mathcal{I}_j$, $\mathcal{I}_q$), we get four absolute rotation predictions $\mathtt{R}_i \mathtt{R}_{\mathcal{I}_i}$, $\mathtt{R}'_i \mathtt{R}_{\mathcal{I}_i}$, $\mathtt{R}_j \mathtt{R}_{\mathcal{I}_j}$, $\mathtt{R}'_j \mathtt{R}_{\mathcal{I}_j}$. In theory, two of them will be identical, while the others differ largely from each other, since $\mathtt{R}_i$ and $\mathtt{R}'_i$ (also $\mathtt{R}_j$ and $\mathtt{R}'_j$) are related by a $180^\circ$ rotation. In practice, we consider the relative rotations (from each pair) that corresponds the two absolute pose predictions with smallest angle difference to be true ones. 

\subsection{Absolute Pose Estimation via RANSAC} 
\label{sec:localization:ransac}
Consider a pair (($\mathcal{I}_i$, $\mathcal{I}_q$), ($\mathcal{I}_j$, $\mathcal{I}_q$)) of image pairs. 
Let $\mathtt{R}_q$ be the absolute rotation of the query image estimated from the two image pairs as described above. 
Let $\mathtt{R}_i$ and $\mathtt{R}_j$ be the relative rotations consistent with $\mathtt{R}_q$.  
Using the two relative translation directions $\mathbf{t}_i$ and $\mathbf{t}_j$, we can determine the position of the query image via triangulation from the two rays $\mathbf{c}_{\mathcal{I}_i} + \lambda_i \mathtt{R}_{\mathcal{I}_i}^T\mathtt{R}_i \mathbf{t}_i$ and $\mathbf{c}_{\mathcal{I}_j} + \lambda_j \mathtt{R}_{\mathcal{I}_j}^T\mathtt{R}_j \mathbf{t}_j$, where $\lambda_i$, $\lambda_j \in \mathbb{R}$ define point positions along the rays. 
The result of triangulation is only defined if the three camera centers are not collinear.  In practice, we use more than two database images to compute the final pose. 
Hence, this will only become a problem in scenarios where all images are taken exactly on a line, \eg, a self-driving car is driving exactly the same route on a line. 

As shown above, a hypothesis for the absolute pose of a query image can be estimated from two pairs. 
To be robust to outlier pairs, we use RANSAC~\cite{Fischler81CACM}: 
In each iteration, we sample two pairs  ($\mathcal{I}_i$, $\mathcal{I}_q$), ($\mathcal{I}_j$, $\mathcal{I}_q$) and use them to estimate the absolute pose hypothesis ($\mathtt{R}_{\mathcal{I}_q}$, $\mathbf{t}_{\mathcal{I}_q}$) of the query image.
Next, we determine which image pairs are inliers to that pose hypothesis. 
For a pair ($\mathcal{I}_k$, $\mathcal{I}_q$) defining four relative poses between $\mathcal{I}_k$ and $\mathcal{I}_q$, we first determine the relative rotation $\mathtt{R}_k$ that minimizes the angle between the absolute rotations $\mathtt{R}_{\mathcal{I}_q}$ and $\mathtt{R}_k\mathtt{R}_{\mathcal{I}_k}$. 
We then use $\mathtt{R}_k$ to measure how consistent the predicted relative pose is with the absolute pose hypothesis predicted by the image pair.  
The relative translation from $\mathcal{I}_q$ to $\mathcal{I}_k$ predicted by the pose ($\mathtt{R}_{\mathcal{I}_q}$, $\mathbf{t}_{\mathcal{I}_q}$) is given as $\mathbf{t}_\text{pred} = \mathtt{R}_{\mathcal{I}_k}(\mathbf{c}_{\mathcal{I}_q} - \mathbf{c}_{\mathcal{I}_k})$. 

We measure the consistency with the predicted relative translation $\mathbf{t}_k$ via the angle 
$\alpha = \cos^{-1}(\mathbf{t}_k^T\mathbf{t}_\text{pred} / (||\mathbf{t}_k||_2||\mathbf{t}_\text{pred}||_2)$. 
If the angle between the two predicted translation directions is below a given threshold $\alpha_\text{max}$, we consider the pair ($\mathcal{I}_k$, $\mathcal{I}_q$) as an inlier.

We use local optimization~\cite{Lebeda2012BMVC} inside RANSAC to better account for noisy relative pose predictions. 
RANSAC finally returns the pose with the largest number of inliers.

\section{ESSENTIAL MATRIX ESTIMATION}
\label{sec:essential_estimation}
While Deep Learning has made huge advances in other vision tasks such as image classification, in visual localiation end-to-end trained methods are still far behind classic methods in terms of accuracy \cite{Sattler2019CVPR}. 
We are highly interested in understanding how to better leverage the power of data-driven methods to build a robust, scalable, flexible and generalizeable localization pipeline. 
To this end, we propose different approaches for essential matrix estimation, ranging from purely hand-crafted to purely data-driven models.
%

\subsection{Feature-based: SIFT + 5-Point Solver}
\label{sec:essential_estimation:sift}
Assuming known camera intrinsics, a classical approach uses local features (in our case SIFT~\cite{Lowe04IJCV}) to establish 2D-2D matches between a query and a database image. 
These matches are then used to estimate the essential matrix by applying the 5-point solver~\cite{Nister2003CVPR} inside a RANSAC loop. 
This approach, which does not need a 3D scene model, serves as a baseline 
within our localization pipeline (\cf Sec.~\ref{sec:localization}). 

\subsection{Learning-based: Direct Regression via EssNet}
\label{sec:essential_estimation:essnet}
The modern alternative to the classical pipeline is to train a CNN for relative pose regression. In the following, we introduce our approach based on essential matrices and discuss its advantages over existing methods. 

\PAR{Relative pose parametrization.} 
Inspired by work on absolute pose regression, \cite{Melekhov2017ICAC, Laskar2017ICCVW, Balntas2018ECCV} propose to directly regress the relative poses with siamese neural networks. ~\cite{Melekhov2017ICAC, Laskar2017ICCVW} parametrize the pose via a rotation and a translation and use the following weighted loss during training 
\begin{equation}\label{eq:weightedloss}
\mathcal{L}_{w}(y^*,\, y)=\left\Vert \mathbf{t}- \mathbf{t}^*\right\Vert_{2}+\beta\left\Vert \mathbf{q} - \mathbf{q}^*\right\Vert_{2} \enspace .
\end{equation}
Here, $y^*=(\mathbf{q}^*\, ,\, \mathbf{t}^*)$ is the relative pose label, $y=(\mathbf{q}\, ,\, \mathbf{t})$ is the relative pose prediction, $\mathbf{q}$ is the relative rotation encoded in a 4D  quaternion, and $\mathbf{t}$ is the 3D translation. 
Notice, the $\beta$ in $\mathcal{L}_{w}$ is a hyperparameter to balance the learning between translation and rotation, which is scene-dependent (\eg, its values differ significantly for indoor and outdoor scenes~\cite{Kendall2015ICCV}).
~\cite{Melekhov2017ICAC} performs grid search to find the optimal $\beta$, following other absolute pose methods~\cite{Kendall2015ICCV, Kendall2016ICRA, Walch2017ICCV}.
\cite{Laskar2017ICCVW} note that setting $\beta=1.0$ works well for indoor scenes. 
Yet, they do not provide any results for outdoor scenes, where finding a single suitable weighting factor is harder due to larger variations in the distance of the camera to the scene. 
\cite{Kendall2017CVPR} propose to learn the weighting parameter  $\beta$ from training data, but are also restricted to a single parameter.

The need for the hyperparameter $\beta$ arises as the rotation (in degrees) and translation (in meters) are typically in different units. 
We note that it can be eliminated through regressing essential matrices, which implicitly define a weighting between an orthonormal rotation matrix and a unit-norm translation vector. 
Tab.~\ref{tab:exp_compare_all} shows that our method based on essential matrix outperforms \cite{Laskar2017ICCVW}, verifying our approach. 

\PAR{Network architecture.} 
We use a siamese neural network 
based on ResNet34~\cite{Kaiming2016CVPR} (until the last average pooling layer) as our backbone (as in ~\cite{Laskar2017ICCVW,Samarth2018CVPR}). 
While~\cite{Melekhov2017ICAC, Laskar2017ICCVW} directly regress relative poses from the concatenated feature maps using with the weighted loss function defined in Eq.~\ref{eq:weightedloss}, we first involve a {\it feature matching} step that resembles the process in classic feature-based localization methods.  
%
We analyze two options for the matching step: 1) a simple fixed matching layer~\cite{Rocco2017CVPR} (\textbf{EssNet}), essentially a matrix dot product between feature maps coming from the two images. 2) a learnable Neighbordhood Consensus (NC) matching layer~\cite{Rocco2018NIPS} (\textbf{NC-EssNet}), which enforces local geometric consistency on the matches. 
%
%
%
Both matching versions combine the two feature maps produced by ResNet into a single feature tensor that can be seen as a matching score map. 

The score map is fed to regression layers to predict the essential matrix. 
The regression layers consist of two blocks of convolutional layers followed by batch normalization with ReLU, and finally a fully connected layer to regress a 9D vector which approximates the essential matrix. 
%
%
%
%
This approximation is then projected to a valid essential matrix by 
replacing the first two singular values of the approximation with their mean value and finally sets the smallest singular value to 0. We use standard functionality provided by PyTorch~\cite{Paszke2017NIPSW} for SVD backpropagation. 

\PAR{Loss function.} During training, we minimize the 
Euclidean distance between the predicted $\mathtt{E}$ and the ground truth essential matrix $\mathtt{E}^*$: 
\begin{equation}\label{eq:essloss}
\mathcal{L}_{ess}(\mathtt{E}^*,\,\mathtt{E})=\left\Vert \mathbf{e}- \mathbf{e}^*\right\Vert_2.
\end{equation}
%
Here, $\mathbf{e} \in \mathbb{R}^9$ is the vectorized $\mathtt{E}\in\mathbb{R}^{3\times3}$.
Given a relative pose label $(\mathtt{R}^*\,,\, \mathbf{t}^*)$, the ground truth essential matrix is $\mathtt{E}^*=[\mathbf{t}^*]_{\times}\,\mathbb{R}^*$, where $[\mathbf{t}^*]_{\times}$ is the skew-symmetric matrix of the normalized translation label $\mathbf{t}^*$, \ie, $||\mathbf{t}^*||=1$. 

\subsection{Hybrid: Learnable Matching + 5-Point Solver}\label{sec:essential_estimation:hybrid}
As a combination of the classical and the regression approaches, we propose a hybrid method: Feature extraction and matching are learned via neural networks, resulting in a set of 2D-2D matches. The 5-point algorithm inside a RANSAC loop is then used to compute the essential matrix. 
In terms of architecture, this approach is equivalent to NC-EssNet without the regression layers.


\section{EXPERIMENTS}\label{sec:experiments}
In the following, we evaluate our novel localization approach based on essential matrix estimation. 
In particular, we are interested in using our approach to analyze why methods based on relative pose regression do not generalize as theoretically expected. 
To this end, we first demonstrate that our approach, based on handcrafted features and RANSAC-based essential matrix estimation, achieves state-of-the-art performance (\cf Sec.~\ref{sec:experiments:sift_5pt}). 
We then use learned essential matrix estimation approaches inside our framework to analyze their weaknesses (\cf Sec.~\ref{sec:experiments:rpr_generalization}). 
Finally, Sec.~\ref{sec:experiments:discussion} discusses our results and draws conclusions for future work.

\PAR{Datasets and evaluation protocol.} 
We follow common practice and use the Cambridge Landmarks~\cite{Kendall2015ICCV} and 7 Scenes~\cite{Shotton2013CVPR} datasets for evaluation. 
For both datasets and all methods, we report the median absolute position error in meters and the median absolute rotation error in degrees, averaged over all scenes within the dataset. 

\PAR{Implementation details.}
We split 1/6 of the training set images as validation images to control the training process. 
Training pairs are generated through image retrieval using the 
CNN (resnet101-gem) proposed in~\cite{Radenovi2018TPAMI}. 

EssNet and NC-EssNet are trained with exact the same settings for fair comparison.
We use a ResNet34 pre-trained on ImageNet~\cite{Deng2009CVPR}. 
The regression network layers are initialized with Kaiming initialization \cite{Kaiming2015ICCV}.  
For each dataset, we train the model on training pairs from {\it all scenes} and evaluate per scene at test time. 
Note, that we use a single network to test on all Cambridge Landmarks sequences, while absolute pose methods~\cite{Kendall2015ICCV,Walch2017ICCV,Kendall2017CVPR,Naseer2017IROS} train a separate network per sequence.
%
All training images are first rescaled so that the shorter side has 480 pixels and then random cropped for training and center cropped for testing to $448\times 448$ pixels. 
All models are trained using the AdamOptimizer\cite{Diederik2015ICLR} with learning rate $1e^{-4}$ and weight decay $1e^{-6}$ in a batch size of 16 for at most 200 epochs.
We early stop training if overfitting is observed and use the model with best validation accuracy.
The code is implemented using Pytorch~\cite{Paszke2017NIPSW} 
and executed on NVidia TITAN Xp GPUs. 

During testing, we use DenseVLAD~\cite{Torii15CVPR} to identify the top-5 most similar training images for each query. 
The retrieved images have to satisfy the following condition designed to avoid retrieving close-by views and thus acute triangulation angles: 
Starting from the top-ranked image, we select the next image that has a distance within  $[a,b]$ meters to all previously selected images. 
For outdoor scenes $a=3, b=50$ and for indoor scenes $a=0.05, b=10$.
We show the choice of RANSAC thresholds $t1$ in the 5-point algorithm~\cite{Nister2003CVPR} to distinguish inliers and outliers and $t2$ in our RANSAC algorithm (\cf Sec.~\ref{sec:localization:ransac}) to remove outlier pairs for absolute pose estimation in Tab.~\ref{tab:exp_ransac_config}. 
The thresholds were chosen through grid-search. 


\begin{table}
    \centering
     \caption{Ransac thresholds used in our experiments.}
 \vspace{-7pt}
\label{tab:exp_ransac_config}
\resizebox{8cm}{!} {
  \begin{tabular}{l c c c}
    \toprule
    Method/Scenes  & (NC-)EssNet & SIFT+5Pt & Learnable Matching +5Pt\\
     \midrule
    Indoor(t1/t2) & -/5 & 0.5/15 & 5.5/20 \\
    Outdoor(t1/t2)& -/5 & 0.5/5 & 4.0/15 \\     
    \bottomrule
 \end{tabular}
 }
 \vspace{-10pt} 
\end{table}


\subsection{Comparison with State-of-the-Art}\label{sec:experiments:sift_5pt}
To validate our pipeline based on essential matrices, we compare results obtained when using SIFT features and the 5-point solver for estimating the essential matrices (\cf Sec.~\ref{sec:essential_estimation:sift}) to state-of-the-art methods. 
We use COLMAP~\cite{Schoenberger2016CVPR} to extract and match features and the 5-point RANSAC implementation provided in OpenCV~\cite{OpenCV}.
%
We compare our approach to methods for absolute pose regression (APR)~\cite{Kendall2015ICCV,Walch2017ICCV,Kendall2017CVPR,Samarth2018CVPR,Kendall2016ICRA}, relative pose regression (RPR)~\cite{Laskar2017ICCVW,Balntas2018ECCV,Saha2018BMVC}, the two image retrieval (IR) baselines based on DenseVLAD~\cite{Torii15CVPR}\footnote{DenseVLAD + Inter. denotes interpolating between the top-ranked database images. See~\cite{Sattler2019CVPR} for details.} used in~\cite{Sattler2019CVPR}, and two state-of-the-art structure-based methods (3D) that explicitly estimate 2D-3D matches~\cite{Brachmann2018CVPR,Sattler2017PAMI}. 
For two RPR methods, we report results obtained when training on the 7 Scenes dataset (7S) and when training on an unrelated dataset (University (U)~\cite{Laskar2017ICCVW} or ScanNet (SN)~\cite{Dai2017CVPR}). 

Tab.~\ref{tab:exp_compare_all} shows that our approach (SIFT+5Pt) consistently outperforms all IR, APR and RPR methods, validating the effectiveness of our pipeline.
Compared to structure-based methods (3D), our approach performs competitively when taking into account that both Active Search and DSAC++ need to build a scene-specific model. 
In contrast, our approach just operates on posed images without the need for using any 3D structure. 
Note that DSAC++ requires two or more days of training while our approach is light-weight and does not require any training.

\begin{table}
\caption{Results on Cambridge Landmarks~\cite{Kendall2015ICCV} and 7 Scenes~\cite{Shotton2013CVPR}.
We report the average median position [m] / orientation [$^\circ$] errors. 
Methods marked with a * build a scene-specific representation and do not generalize to unseen scenes.}
 \vspace{-7pt}
\label{tab:exp_compare_all}
\scriptsize{
\setlength{\tabcolsep}{4pt}
\begin{center}
\begin{tabular}{cl|c|c}
& & Cambridge Landmarks & 7 Scenes \\
\multirow{2}*{\begin{sideways} IR \end{sideways}} 
& DenseVLAD~\cite{Torii15CVPR} & \hco{2.56/7.12} & \hco{0.26/13.11}\\
& DenseVLAD + Inter.~\cite{Sattler2019CVPR} & \hco{1.67/4.87} & \hco{0.24/11.72} \\ \hline \hline 

\multirow{2}*{\begin{sideways} 3D \end{sideways}} 
& *Active Search~\cite{Sattler2017PAMI} & 0.29/0.63  & 0.05/2.46\\
& *DSAC++~\cite{Brachmann2018CVPR} & \textbf{0.14/0.33} & \textbf{0.04/1.10}\\
\hline\hline 
\multirow{7}*{\begin{sideways} APR \end{sideways}}
& *PoseNet (PN)~\cite{Kendall2015ICCV} & \hco{2.09/6.84} & \hco{0.44/10.44} \\
& *Learn. PN~\cite{Kendall2017CVPR} & \hco{1.43/2.85} & \hco{0.24/7.87} \\
& *Bay. PN~\cite{Kendall2016ICRA} &\hco{1.92/6.28} & \hco{0.47/9.81}\\
& *Geo. PN~\cite{Kendall2017CVPR} & \hco{1.63/2.86} & \hco{0.23/8.12}\\
& *LSTM PN~\cite{Walch2017ICCV} & \hco{1.30/5.52} & \hco{0.31/9.85}\\
& *MapNet~\cite{Samarth2018CVPR} & \hco{1.63/3.64} & 0.21/\hco{7.78} \\
& *MapNet+PGO~\cite{Samarth2018CVPR} & - & 0.18/6.56\\ 
\hline
\multirow{5}*{\begin{sideways} RPR \end{sideways}} 
& Relative PN~\cite{Laskar2017ICCVW} (U) & - & \hco{0.36/18.37} \\
& Relative PN~\cite{Laskar2017ICCVW} (7S) & - & 0.21/\hco{9.28}\\
& RelocNet~\cite{Balntas2018ECCV} (SN) & - & \hco{0.29/11.29}\\
& RelocNet~\cite{Balntas2018ECCV} (7S) & - & 0.21/6.73\\ 
& *AnchorNet~\cite{Saha2018BMVC} & 0.84/2.10 & 0.09/6.74\\  
\hline\hline 
\multirow{7}*{\begin{sideways} \textbf{Ours} \end{sideways}} 
& Sift+5Pt & 0.47/0.88 & 0.08/1.99\\ \cline{2-4} 
& EssNet   & 1.08/3.41 & 0.22/8.03\\  
& \textbf{\hco{NC-EssNet}} & 0.85/2.82 & 0.21/7.50\\ \cline{2-4} 
& NC-EssNet(7S)+NCM+5Pt & 0.89/1.39 & 0.19/4.28\\ 
& Imagenet+NCM+5Pt & 0.83/1.36 & 0.19/4.30\\ 
& EssNet224(SN)+NCM+5Pt &  0.90/1.37& 0.19/4.35\\ %
\hline
\end{tabular}
\end{center}
}%
\vspace{-9pt}
\end{table}

\subsection{Analyzing Relative Pose Regression (RPR)}
\label{sec:experiments:rpr_generalization}
One motivation for our localization pipeline is to understand why RPR methods perform worse compared  to structure-based methods. 
In the following experiment, we use (NC-)EssNet (\cf Sec.~\ref{sec:essential_estimation:essnet}) as the RPR method inside our pipeline.

\PAR{Comparison with state-of-the-art.} 
%
%
Tab.~\ref{tab:exp_compare_all} compares our approaches against the current state-of-the-art. For visibility, we mark results that are less accurate than NC-EssNet in \hco{red}.
As can be seen, NC-EssNet, our best regression model, performs better than all APR approaches except for MapNet+PGO which uses external GPS information. 
Also, our NC-EssNet is competitive to RelocNet and outperforms Relative PN.
While NC-EssNet is less accurate than AnchorNet~\cite{Saha2018BMVC}, AnchorNet  
needs to be trained explicitly per scene as it encodes training images in the network.  
The results show that our methods achieve state-of-the-art performance among pose regression methods. 

\begin{table}
 \caption{Generalization study of regression models. We show average median position (in meters) / orientation (in degrees) errors.}
  \vspace{-7pt}
\label{tab:exp_generalization}
    \centering
\resizebox{7cm}{!} {
  \begin{tabular}{l c c r}
    \toprule
     Essential Matrix & Training & \multicolumn{2}{c}{Testing Data}\\
    Estimation &  Data  & Cambridge & 7Scenes \\
     \midrule
	 EssNet		& Cambridge & 1.08/3.41 & \hcp{0.57/80.06} \\ 
	 NC-EssNet	& Cambridge & 0.85/2.82 & \hcp{0.48/32.97} \\ 
     EssNet		& 7Scenes & \hcp{10.36/85.75} & 0.22/8.03 \\ 
     NC-EssNet	& 7Scenes & \hcp{7.98/24.35} & 0.21/7.50 \\ 
	 SIFT+5Pt	& -  &  0.47/0.88 & 0.08/1.99 \\	
    \bottomrule
 \end{tabular}
 }
    \vspace{-7pt}
\end{table}

\PAR{Failure to generalize.} 
Compared to absolute pose regression, the promise of relative pose regression is generalization to new scenes~\cite{Laskar2017ICCVW,Balntas2018ECCV}: 
An absolute pose estimate is scene-specific as it depends on the coordinate system used. 
In contrast, a network that learns to regress a pose relative to another image could learn general principles that are applicable to unseen scenes. 

Tab.~\ref{tab:exp_generalization} analyzes the ability of EssNet and NC-EssNet to generalize from indoor to outdoor scenes and vice versa, where we mark failure cases in \hcp{purple}. 
As can be seen, there is a substantial gap in pose accuracy compared to training on the same scenes and especially compared to the classical variant (SIFT+5Pt) of our pipeline. 
This clearly indicates that EssNet and NC-EssNet fail to learn a general underlying principle. 
As similar observation holds for~\cite{Laskar2017ICCVW,Balntas2018ECCV} in Tab.~\ref{tab:exp_compare_all}, based on the performance when trained on 7 Scenes (7S) and on another  dataset (U or SN). 

Looking at Tab.~\ref{tab:exp_generalization} and Tab.~\ref{tab:exp_compare_all}, the important question to ask is why RPR methods fail to generalize:  
Do the features extracted in their base networks fail to generalize, is there a lack of generalization in the layer that regresses the relative pose, or is it a combination of both? 
In order to better understand the behavior of EssNet and NC-EssNet, we consider the hybrid version of our pipeline (\cf Sec.~\ref{sec:essential_estimation:hybrid}).

The hybrid variant \textit{always} uses the NC matching layer (NCM) trained on the unrelated ivd dataset~\cite{Rocco2018NIPS} to extract feature matches.
%
%
To analyze the impact of the \textbf{feature extraction} on the generalization performance, we compare our \textit{ResNet34} backbones trained in different ways and on multiple datasets, \eg, the pretrained model for the image classification (IC) task on ImageNet~\cite{Russakovsky2015IJCV} and our trained models for the essential matrix regression (EMR) task on MegaDepth(MD)~\cite{Li18CVPR} (outdoor), 7 Scenes(7S)~\cite{Shotton2013CVPR} (indoor), and Cambridge Landmarks(CL)~\cite{Kendall2015ICCV} (outdoor) datasets.
In order to make training computationally feasible on large datasets such as MegaDepth and ScanNet, we train EssNet with reduced image resolution ($224\times 224$). 
We denote these trained \textit{feature extractors} with EssNet224.
%
Note that for our hybrid, we perform inference with the original high resolution images. 


Tab.~\ref{tab:exp_feature_compar} evaluates the performance of the different training strategies on the localization accuracy of our hybrid approach. 
As can be seen, there is little variation in performance independently how the features are trained. 
This clearly shows that the features themselves generalize well and that the failure to generalize observed in Tab.~\ref{tab:exp_generalization} is caused by the regression layers. 

\begin{table}
    \centering
     \caption{Evaluating the impact of trained features on localization performance when using different training strategies.}
\label{tab:exp_feature_compar}
\resizebox{8.5cm}{!} {
  \begin{tabular}{l c c c r}
    \toprule
    Feature Matching & Train Task & Train Data& Cambridge & 7Scenes \\
     \midrule
     ImageNet+NCM& IC & ImageNet~\cite{Russakovsky2015IJCV} & 0.83/1.36 & 0.19/4.30\\ 
	 NC-EssNet+NCM		& EMR & 7S &  0.89/1.39& 0.19/4.28\\  
	 NC-EssNet+NCM		& EMR & CL &  0.96/1.43 & 0.20/4.61 \\  
	 EssNet224+NCM		& EMR & MD~\cite{Li18CVPR} &  0.98/1.4& 0.20/4.70\\  
	 EssNet224+NCM		& EMR & SN~\cite{Dai2017CVPR} &  0.90/1.37& 0.19/4.35\\  
	 EssNet224+NCM		& EMR & MD+7S+CL &  0.96/1.48 & 0.23/4.89 \\  
	 SIFT+5Pt	& -  & - & 0.47/0.88 & 0.08/1.99 \\	
    \bottomrule
 \end{tabular}
 }
\end{table}

\subsection{Discussion}
\label{sec:experiments:discussion}
In a classical approach~\cite{Hartley2004}, the relative pose between two images is estimated by finding feature correspondences in the image pair. 
%
%
When directly regressing the relative pose/essential matrix from an image pair, we can only assume that an implicit feature matching is performed within regression. In contrast, our hybrid approach explicitly learns the feature matching task and adopts the established multi-view geometry knowledge to compute relative poses from correspondences.
The fact that the relative pose regression layers fail to generalize to unseen scenes and to produce accurate poses implies 
that the implicit matching cannot be properly learned by a regression network.
While explicitly learning the matching task leads to better generalization, the resulting poses are still not as accurate as the poses estimated by SIFT+5pt, as can be seen in Tab.~\ref{tab:exp_feature_compar}.
Such inaccuracy is related to the fact that the current CNN features are coarsely localized on the images, that is, the features from later layers are not mapped to a single pixel but rather an image patch. One possible solution would be networks designed to obtain better localized features\cite{Dusmanu2019CVPR, Detone2018CVPRW}.
Another would be to follow ~\cite{Ranftl2018ECCV,Yi2018CVPR,Dang2018ECCV}, where a network is trained to detect outliers, and can be applied 
as a post-processing step for any type of matches.
However, integrating those methods into an end-to-end pipeline going from image pairs to poses is not straight-forward and will constitute interesting future work. 

\section{CONCLUSION}
In this paper, we have proposed a novel framework for visual localization from essential matrices.  
Our approach is light-weight and flexible in the sense that it does not use information about the 3D scene structure model of the scene and can thus easily be applied to new scenes. 
Our results show that our framework can achieve state-of-the-art results. 
We have evaluated our framework using three different methods for computing essential matrices, ranging from purely hand-crafted to purely data-driven. 
By comparing their results, we have shown that the purely data-driven approach does not generalize well and have identified the reason for this failure as the relative pose regression layers. 
Furthermore, we have shown that the features and matches used by the data-driven approach themselves generalize quite well. 
However, directly using them for pose estimation yields less accurate results compared to the hand-crafted version of our pipeline. 
%
Based on our analysis, it is clear that more research is required before data-driven visual localization methods perform accurately and easily generalize to new scenes. 

{\small
\bibliographystyle{ieee}
\bibliography{pose-icra19.bib}
}

\end{document}